\documentclass[10pt,twocolumn,letterpaper]{article}

\usepackage[pagenumbers]{cvpr} 
\usepackage{bm}
\usepackage{multirow}
\usepackage{graphicx}
\usepackage{amsmath}
\usepackage{booktabs}

\usepackage{color}
\usepackage{colortbl}

\usepackage[dvipsnames]{xcolor}

\definecolor{cvprblue}{rgb}{0.21,0.49,0.74}
\definecolor{Gray}{gray}{0.90}
\usepackage[pagebackref,breaklinks,colorlinks,citecolor=cvprblue]{hyperref}

\definecolor{mygray}{rgb}{.95,.95,.95}
\definecolor{myblue}{rgb}{0.94, 0.95, 1.0}

\def\paperID{15745} 
\def\confName{CVPR}
\def\confYear{}

\title{GeoChat\includegraphics[width=0.04\linewidth]{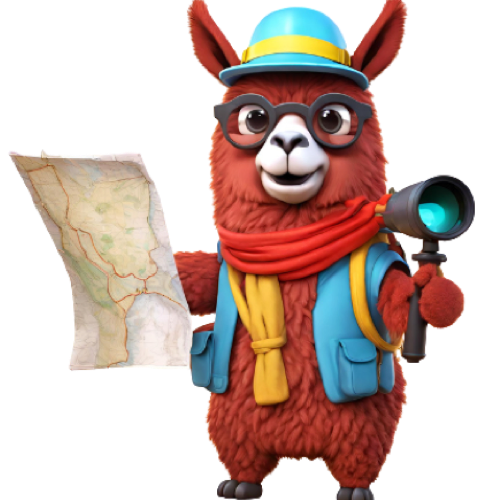}: Grounded Large Vision-Language Model for Remote Sensing}

\author{Kartik Kuckreja\textsuperscript{1, 2}\thanks{Equally contributing first authors.}
\and
Muhammad Sohail Danish\textsuperscript{1*}
\and
Muzammal Naseer\textsuperscript{1} 
\and 
Abhijit Das\textsuperscript{2} \qquad\qquad
Salman Khan\textsuperscript{1, 3}
\and
Fahad Shahbaz Khan\textsuperscript{1, 4}
\and
\small{
\textsuperscript{1}Mohamed bin Zayed University of AI,  \textsuperscript{2}Birla Institute of Technology \& Science, Hyderabad} \\
\small{\textsuperscript{3}Australian National University, \textsuperscript{4}Linköping University }\\
{\tt\small \href{Kartik.Kuckreja@mbzuai.ac.ae}{kartik.kuckreja@mbzuai.ac.ae}, \href{muhammad.sohail@mbzuai.ac.ae}{muhammad.sohail@mbzuai.ac.ae} }
}

\begin{document}
\maketitle
\begin{abstract}
\vspace{-2em}
Recent advancements in Large Vision-Language Models (VLMs) have shown great promise in natural image domains, allowing users to hold a dialogue about given visual content. 
However, such general-domain VLMs perform poorly for Remote Sensing (RS) scenarios, leading to inaccurate or fabricated information when presented with RS domain-specific queries. 
 Such a behavior emerges due to the unique challenges introduced by RS imagery. For example, to handle high-resolution RS imagery with diverse scale changes across categories and many small objects, region-level reasoning is necessary alongside holistic scene interpretation. Furthermore, the lack of domain-specific multimodal instruction following data as well as strong backbone models for RS make it hard for the models to align their behavior with user queries. To address these limitations, we propose GeoChat - the first versatile remote sensing VLM that offers multitask conversational capabilities with high-resolution RS images. Specifically, GeoChat can not only answer image-level queries but also accepts region inputs to hold region-specific dialogue. Furthermore, it can visually ground objects in its responses by referring to their spatial coordinates. To address the lack of domain-specific datasets, we generate a novel RS multimodal instruction-following dataset by extending image-text pairs from existing diverse RS datasets.
 We establish a comprehensive benchmark for RS multitask conversations and compare with a number of baseline methods. GeoChat demonstrates robust zero-shot performance on various RS tasks, e.g., image and region captioning, visual question answering, scene classification, visually grounded conversations and referring detection. Our code is available \href{https://github.com/mbzuai-oryx/geochat}{here}.
 
\end{abstract}

\section{Introduction}
\vspace{-.5em}
\label{sec:intro}

\begin{figure}
	\begin{center}             
        \includegraphics[width=1\linewidth]{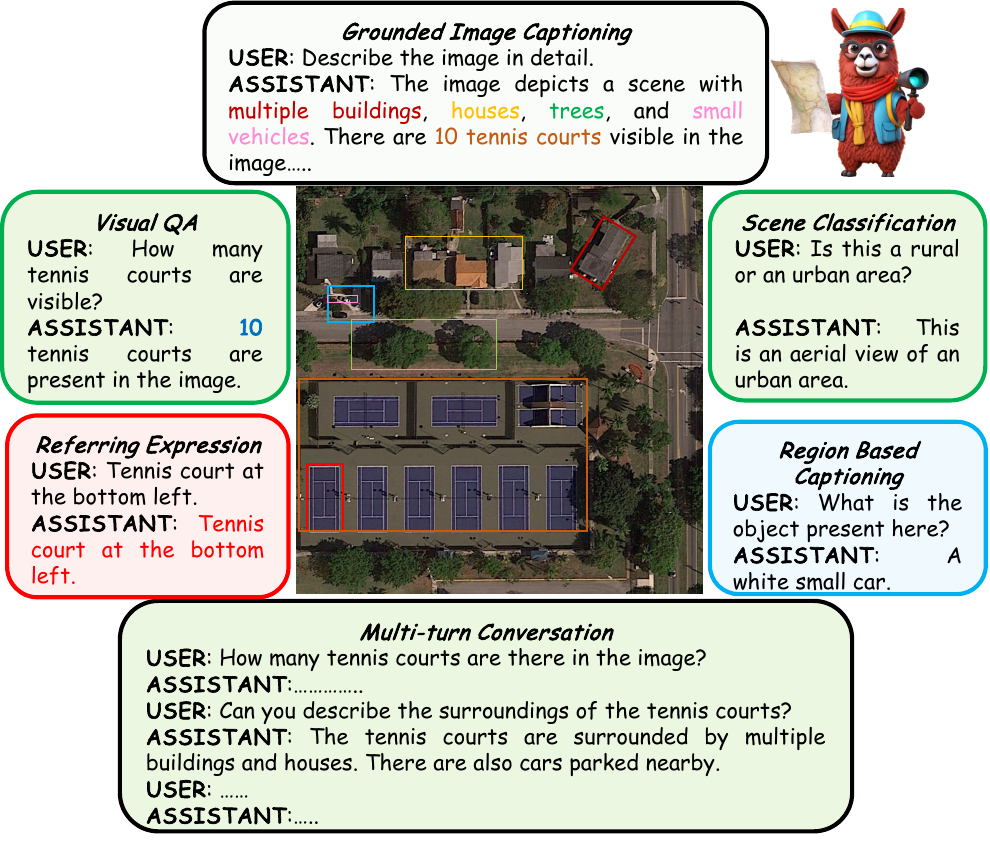}
    \end{center}
\vspace{-0.2cm}
    \caption{GeoChat can accomplish multiple tasks for remote-sensing (RS) image comprehension in a unified framework. Given suitable task tokens and user queries, the model can generate visually grounded responses (text with corresponding object locations - shown on top), visual question answering on images and regions (top left and bottom right, respectively) as well as scene classification (top right) and normal natural language conversations (bottom). This makes it the first RS VLM with grounding capability.  }
    \label{fig:teaser}          
\end{figure}

In the natural image domain, the abundance of aligned image-text data sourced from web imagery or manual annotations facilitate effective self-supervised vision-language modeling, as demonstrated by multimodal GPT-4~\cite{openai2023gpt4} and open-source initiatives like LLaVA~\cite{llava}. 
These vision-language models (VLMs), developed through generative pretraining and instruction-tuning, exhibit robust zero-shot task completion across various user-oriented multimodal tasks. 
The resulting capabilities open the door to the development of versatile multimodal conversational assistants with broad applications in real-world scenarios \cite{hu2023rsgpt}.

However, general-domain VLMs designed for natural images, exhibit poor performance when presented with remotely sensed visual imagery. 
The performance disparity arises primarily from the distinct nature of content found in remote sensing image-text pairings compared to the publicly available web data. 
As a result, general-domain VLMs can provide inaccurate information or hallucinate when presented with spatial images from RS sensors. 
Although there has been significant progress in the field of remote sensing visual question answering (VQA)~\cite{yuan2022easy, zhang2023spatial}, earlier methods have framed the task as a classification problem. Here, the model chooses answers from predetermined responses found in the training data. It limits their applicability to open-ended answer generation and instruction-following.

In this paper, we introduce {GeoChat}, an attempt to extend multimodal instruction-tuning to the remote sensing domain for training a multitask conversational assistant. However, remote-sensing domain lacks a multimodal instruction-tuning conversational dataset. 
Inspired by recent work in instruction-tuning \cite{llava,li2023llava,minigpt}, GeoChat uses Vicuna-v1.5 \cite{vicuna2023} and an automated pipeline to generate diverse remote sensing multimodal instruction-following data comprising of nearly $318k$ instructions.  We create the image-text pairs from various existing remote sensing datasets developed for diverse tasks. These includes LRBEN for VQA \cite{lobry2020rsvqa}, NWPU-RESISC-45 for scene classification \cite{cheng2017remote} and SAMRS for object detection \cite{SAMRS}.  

A crucial capability of GeoChat is the unification of multiple image and region-level reasoning tasks for RS imagery within a single pipeline (see Fig.~\ref{fig:teaser}). We achieve this via distinct task tokens that help suitably direct the model's responses according to user requirements. In addition, the model uses spatial location representations in its inputs to seamlessly reason about local regions and can also generate object locations in its responses to visually ground objects. This enables a diverse set of tasks possible with GeoChat including referring expression detection, image/region captioning, scene classification, natural language conversations and VQA, besides visually grounded conversations.


In summary, this work has the following contributions:
\begin{itemize}
  \item \textit{RS multimodal instruction following dataset}. We present a novel data generation pipeline, to leverage existing object detection dataset \cite{SAMRS} to create short descriptions of the images, followed by using Vicuna-v1.5 \cite{vicuna2023} to create conversations using the generated text alone. Further, we add visual question-answering and scene classification abilities 
 using their corresponding datasets \cite{lobry2020rsvqa,cheng2017remote}. This results in a total of $318k$ instruction pairs for RS domain.

  \item \textit{GeoChat}. Leveraging our dataset, we finetune LLaVA-1.5 \cite{li2023llava} to create the remote sensing-domain vision-language model - GeoChat. Our LoRA \cite{lora} fine-tuning is efficient and avoids forgetting the necessary context embedded in fully-tuned LLaVA model, whose MLP projection is trained to align images into the word embedding space of the LLM (Vicuna-v1.5 \cite{vicuna2023}). This allows GeoChat to retain the conversation and instruction following abilities of LLaVA and extend its domain-knowledge to remote sensing tasks.  

  \item We also address the lack of evaluation benchmarks to assess the capability of existing VLMs on remote-sensing conversations. To this end, we setup evaluation protocols for conversation grounding in RS, as well as a setup a suite of tasks to allow comparisons with future efforts in this direction. We show various supervised as well as  zero-shot evaluations for different remote sensing tasks, including image captioning, visual question answering and scene classification to demonstrate the generalisability of GeoChat conversational VLM.
  
\end{itemize}

\section{Related Work}
\label{sec:relatedwork}

\textbf{Large Vision-Language Models.}
The typical architecture of instruction-following Vision Language Models  (VLMs) consists of utilising a pre-trained visual backbone\cite{vit} to encode visual data, a large language model~\cite{vicuna2023} for interpreting user instructions and generating responses, and a vision-language cross-modal connector, e.g., a linear projection layer \cite{minigpt,liu2023visual} or an MLP \cite{liu2023improved}, for fusing visual information with language models. The results achieved with VLMs show great promise; for example, LLaVA~\cite{liu2023visual}, Instruct-BLIP~\cite{instructblip}, Otter~\cite{li2023otter} and MiniGPT-4~\cite{minigpt} show remarkable gains in language instruction following and visual reasoning ability for natural scenes. More recent studies have shown that these models can be adapted to other domains such as videos~\cite{maaz2023video}, biomedical~\cite{li2023llava,Omkar2023XrayGPT} and remote sensing~\cite{hu2023rsgpt}.

\noindent\textbf{Remote Sensing VLMs.}
The application of generalized VLMs in remote sensing is comparatively sparse. The majority of research so far has neglected the semantic understanding of the items and their relationships towards a deep visual comprehension. Beyond merely identifying the objects in an image, vision-language models are also capable of generating natural language descriptions of the image and inferring the connections between the objects. This makes them more appropriate for tasks like text-based image retrieval, captioning images, and answering visual questions that call for both visual and linguistic knowledge. Although there has been progress in vision language models for remote sensing tasks, such as image captioning \cite{zia2022transforming}, zero-shot classification \cite{li2023rs} and visual question answering \cite{yuan2022easy,chappuis2022prompt}, these models can only perform a specific task they are trained for, lack conversational capability and do not possess generic semantic knowledge about the remote sensing images. A major gap exists in the remote sensing domain towards developing general-purpose models to solve all tasks together, while also maintaining conversation abilities. While RSGPT~\cite{hu2023rsgpt} is an initial effort that has shown good conversation ability along with solving multiple tasks, it requires finetuning the model for each task separately, which makes it cumbersome and not generalizable. Further, RSGPT cannot work for region-level reasoning or visual grounding, which our work aims to address.

\section{GeoChat: Grounded Remote Sensing VLM}


\begin{figure*}
\vspace{-0.2cm}
	\begin{center}             
        \includegraphics[width=1\linewidth]{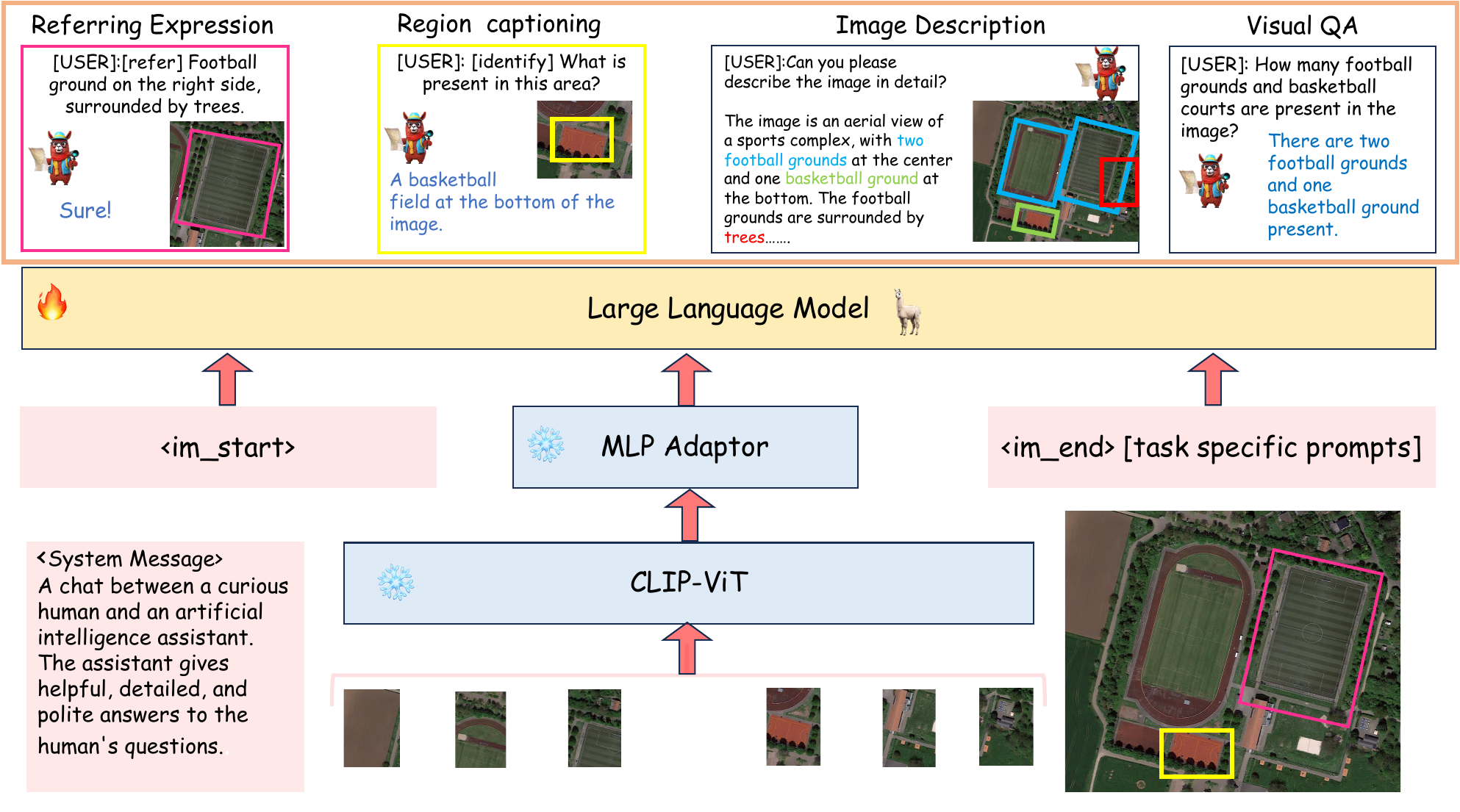}
    \end{center}
\vspace{-0.5cm}

    \caption{An overview of GeoChat - the first grounded large vision-language model for remote sensing. Given an image input together with a user query, a visual backbone is first used to encode patch-level tokens at a higher resolution via interpolating positional encodings. A multi-layer perceptron (MLP) is used to adapt vision-tokens to language space suitable for input to a Large Language Model (Vicuna 1.5). Besides visual inputs, region locations can also be input to the model together with task-specific prompts that specify the desired task required by the user. Given this context, the LLM can generate natural language responses interleaved with corresponding object locations. GeoChat can perform multiple tasks as shown on top e.g., scene classification, image/region captioning, VQA and grounded conversations.}
    \label{fig:graph}          
\end{figure*}

Visually grounded conversations for remote sensing aim to generate textual responses interleaved with corresponding object locations. Further, a user can also provide visual prompts (e.g., a bounding box) besides natural language questions, and the model should be able to answer questions about the specified Region of Interest (RoI). Such seamless interplay between visual and language modalities necessitate a deep comprehension of linguistic constructions that denote particular objects or elements in a visual scene.

As mentioned above, GeoChat is the first model capable of holding visually grounded conversations about remotely sensed images. By construction, GeoChat can address not only the challenging task of visually grounded conversations, but can also perform a  spectrum of other spatial reasoning tasks that span varying levels of granularity in visual imagery understanding e.g., image/region captioning, referring object detection and image/region-level conversations about remotely sensed images. We formally outline the tasks possible with GeoChat below.


\textbf{a) Image-Level Conversation Tasks.} In this task, GeoChat processes an image $\bm{x}$ and a user text query $\bm{q}$ without any specific spatial coordinates in its inputs or outputs. The goal is to perform conversation-based tasks at a holistic level with image-wide context, such as visual question answering (VQA), scene classification and image captioning.

\textbf{b) Region-Level Conversation Tasks.} This task involves providing spatial box locations $\bm{b}$ in the input to GeoChat besides $\bm{x}$ and $\bm{q}$. Region locations $\bm{b}$ guide the model's attention to specific regions within the image, so that the model can perform tasks such as region-level captioning, region-specific VQA or multi-turn conversation.

\textbf{c) Grounded Conversation Tasks.} With the use of special tokens, termed as task-specification tokens $\bm{t}$, GeoChat can be guided to provide object locations at different granularities, while maintaining conversation abilities. It helps in tasks including grounded image captioning/conversation, object grounding and referring expression detection.

\begin{table}[!htp]
  \centering
  \resizebox{\columnwidth}{!}{
  \begin{tabular}{lcc}
    \toprule
   \rowcolor{mygray} Data & Size & Response formatting prompts  \\
    \midrule
    Detailed Description & 30k & Describe the image in detail. \\
    \midrule
     Multi-Round Conversation& 65k & -\\

      Complex Questions & 10k & -\\
      \midrule
     RSVQA-LRBEN\cite{lobry2020rsvqa} & 56k & Answer the question using a single word or phrase. \\
     NWPU-RESISC-45\cite{cheng2017remote} & 31.5k & \\
     Floodnet\cite{rahnemoonfar2020floodnet}& 4k & \\
     \midrule
    Grounding Description & 45k & \texttt{[grounding]}  Describe the image in detail. \\
     \bottomrule
     Region Captioning & 40k & \texttt{[identify]} \( \{b\textsubscript{x\_left}, b\textsubscript{y\_top}, b\textsubscript{x\_right}, b\textsubscript{y\_bottom} | \theta \} \)\\
     \bottomrule
     Referring Expression & 25k & \texttt{[refer]} \(<p>\) Object \(</p>\)\\
    \bottomrule
  \end{tabular}}
    \caption{ Instruction following data used to train GeoChat. Instruction types and format are shown. We use a 306k set for training and a separate 12k instruction-set for testing.  }
    \label{tab:instruction_following_data_mixture}
\end{table}

\subsection{GeoChat Architecture}

GeoChat follows the architecture as of LLaVA-v1.5~\cite{liu2023improved}, which consists of three core components, i) Global Image encoder, ii) an MLP adaptor (two linear layers) and iii) LLM. 
Different to LLaVA, we add specific task prompt that indicates the type of task desired from the model i.e., grounding, image-level or region-level conversations. Additionally, we allow spatial positions within both inputs and outputs, enabling visual prompts as inputs and grounded objects in GeoChat outputs. Notably, the original LLaVA model cannot perform object grounding or accept region inputs. Further, the original LLaVA can not reason about remote sensing images which is enabled via our domain-specific dataset. 
We describe each component in the architecture as follows:

\textbf{Task Token:} The unique quality of GeoChat is its ability to easily switch between different types of remote sensing visual interpretation tasks.
To eliminate uncertainty among tasks, our approach assigns a unique task identification to each one. We suggest three distinct task identities, $\bm{t} \in \,$\texttt{\{grounding, identify, refer\}}, each for grounded conversations, region captioning and referring expression comprehension. As for the case of visual question answering and scene classification, we directly ask the model to output the answer in a single word or phrase, as shown in Table \ref{tab:instruction_following_data_mixture}.
Our approach does not employ any task identification tokens for vision-irrelevant commands. This unified approach is supported by a modular design that efficiently integrates spatial data, giving the model flexibility in its reasoning about visual content. 

\textbf{Spatial Location Representation.} Our model must precisely identify the spatial position of the referenced items for tasks such as grounded conversations, referring expression generation, and comprehension. To this end, we represent the box locations in a textual format to express the geographical position: \(\bm{b} = \{b\textsubscript{x\_left}, b\textsubscript{y\_top}, b\textsubscript{x\_right}, b\textsubscript{y\_bottom} | \theta\}\). Here, \(b\textsubscript{x\_left}, b\textsubscript{y\_top}\) denote the top left corner point of box while the \(b\textsubscript{x\_right}, b\textsubscript{y\_bottom}\) represent the bottom right corner coordinates. The angle \(\theta\) represents the angle of rotation for the bounding box, from the lower edge. Numerical values normalised within the interval [0, 100] are used to represent the x and y coordinates. Region locations in this format are used to interact with the model via its inputs and outputs. 

\textbf{Visual Backbone.} GeoChat adapts the pretrained vision backbone of CLIP-ViT(L-14)~\cite{tay2017learning}, which has an input resolution of 336$\times$336. This results in effectively 576 patches per image. Since this resolution is not sufficient to understand details presented in remote sensing imagery (e.g., small objects and object details), we interpolate the positional encoding in the transformer-based CLIP~\cite{tay2017learning} model to scale with input image sizes of 504$\times$504. Although this leads to an increase in the number of patches to almost double (i.e., 1296 per image), this enhanced resolution allows us to handle larger image sizes and also supports better visual grounding in high-resolution RS images.

\textbf{MLP Cross-modal Adaptor.} From the frozen CLIP-ViT\cite{tay2017learning}, we project the output tokens ($\in \mathbb{R}^{1296\times 1024}$) with dimensions 1024 onto the language model space, using an MLP adaptor with one hidden layer. The adaptor has an input dimensionality of 1024 and outputs a vector of size 4096, corresponding to the input size of the LLM \cite{vicuna2023}. A  GeLU \cite{hendrycks2016gaussian} is used as the activation function.

\textbf{Large Language Model.} The open source Vicuna-v1.5(7B)~\cite{vicuna2023} large language model is utilised as the foundation for GeoChat. The language model functions as a single interface for diverse vision-language inputs in our framework. To accomplish different vision-language tasks, we directly depend on the Vicuna-v1.5(7B)~\cite{vicuna2023} language tokens. We explicitly interact with the language model to construct textual representations of bounding boxes to express their spatial coordinates for the visual grounding tasks that require the production of spatial locations. Similarly, the safe, aligned and effective behavior of LLM is ensured via system prompts appended together with given inputs. 
A  Low-Rank Adaptation (LoRA) \cite{lora} based strategy is used for fine-tuning the LLM.
While training, instead of finetuning all of the weights that comprise the weight matrix of the pre-trained Vicuna-v1.5\cite{vicuna2023}, we finetune two smaller matrices in LoRA~\cite{lora} that approximate the original larger matrix. 
After that, the fine-tuned adaptor is fed into the pretrained model and utilised for inference. The LoRA adaptation ensures faster training and avoids forgetting original knowledge embedded in the LLM trained and fine-tuned on generic natural language instructions. This is an important feature since it allows the model to bring in external context about generic object types, landmarks and affordances in the remote-sensing reasoning framework of GeoChat. 



\begin{figure}
\vspace{-0.2cm}
	\begin{center}             
        \includegraphics[width=1\linewidth]{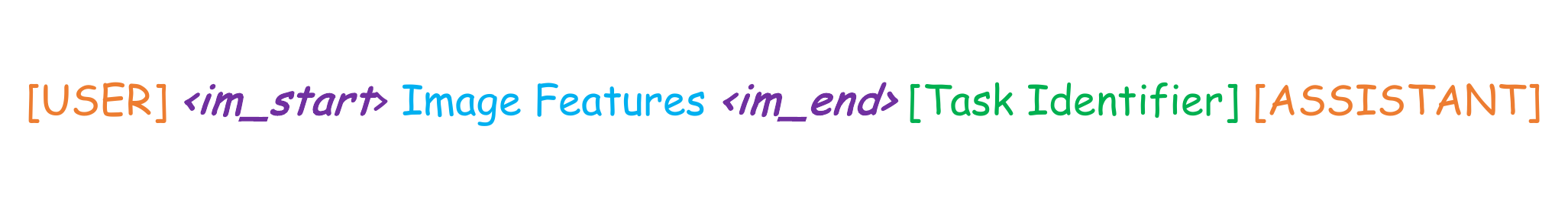}
    \end{center}
\vspace{-0.6cm}
    \caption{ Multi-task instruction template for GeoChat. }
    \label{fig:instr template}          
\end{figure}

\begin{figure*}
\vspace{-0.2cm}
	\begin{center}             
        \includegraphics[width=1\linewidth]{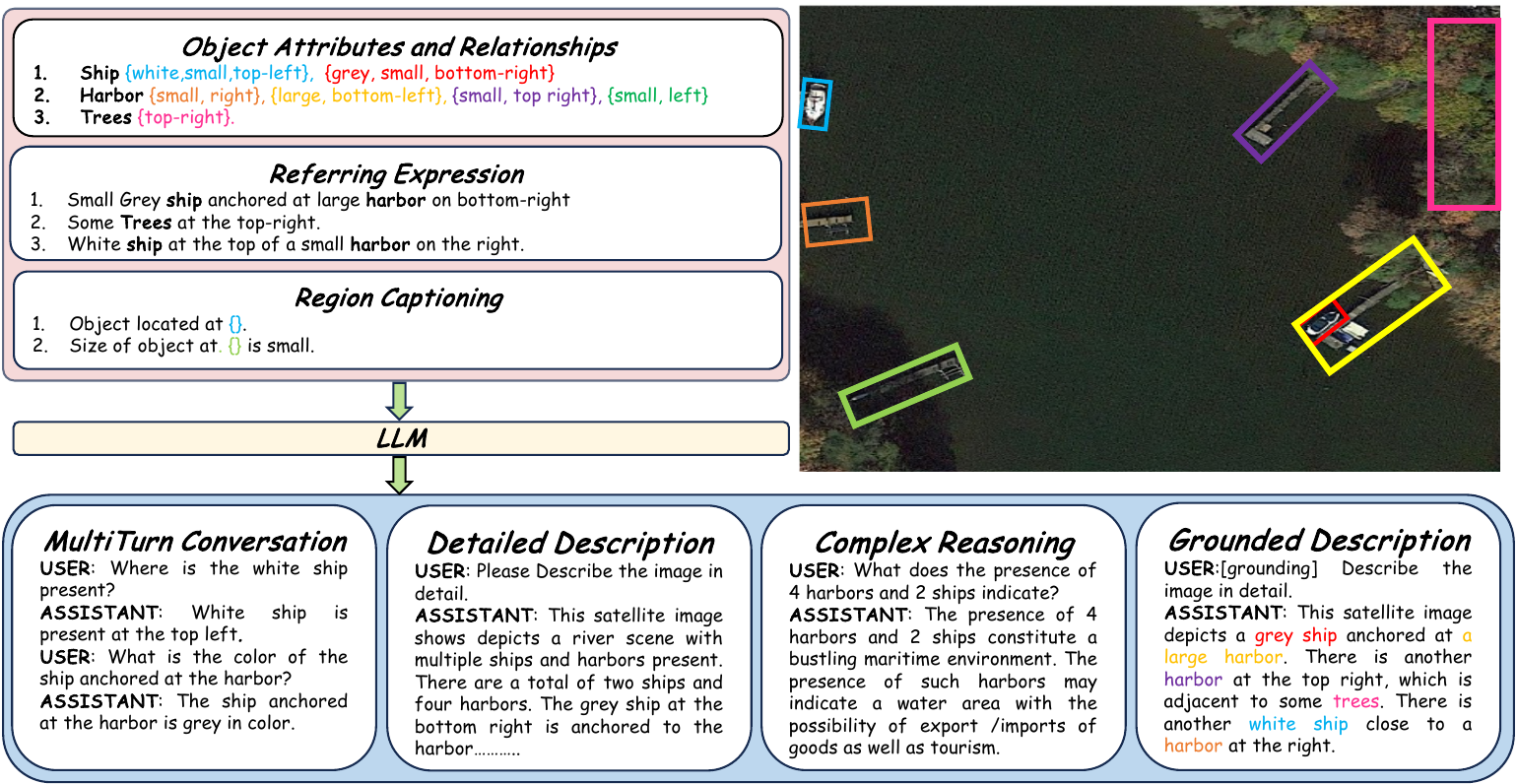}
    \end{center}
\vspace{-0.4cm}
    \caption{Types of annotations available in the GeoChat instruction-set. \emph{Top-row:} For a given RS image, we obtain object attribute and relationship information, referring expressions and region captions along with their corresponding region annotations (shown over the image). \emph{Bottom-row:} This structured information is used to create the rich instruction-set with a total of 318k image-instruction pairs.}
    \label{fig:graph}          
\end{figure*}

\subsection{Training Details}
To enhance the effectiveness of our model on general visual tasks and optimize training efficiency, we employ a strategy that involves initializing the network with pre-trained weights and fine-tuning specific segments for remote sensing related tasks. We use a pre-trained CLIP-ViT(L-14) encoder\cite{tay2017learning},trained on large amounts of textual and visual data, a pretrained MLP adaptor\cite{liu2023improved}, pretrained on a 558K subset of the LAION-CC-SBU \cite{schuhmann2021laion} dataset with BLIP \cite{li2022blip} captions, and Vicuna-v1.5\cite{vicuna2023} to initialize our model. To adapt our model to remote sensing images, we subsequently LoRA \cite{lora} fine-tune the LLM , while keeping the MLP adaptor and the CLIP encoder \cite{tay2017learning} frozen during training.

\begin{figure*}
\vspace{-0.2cm}
	\begin{center}             
        \includegraphics[width=1\linewidth]{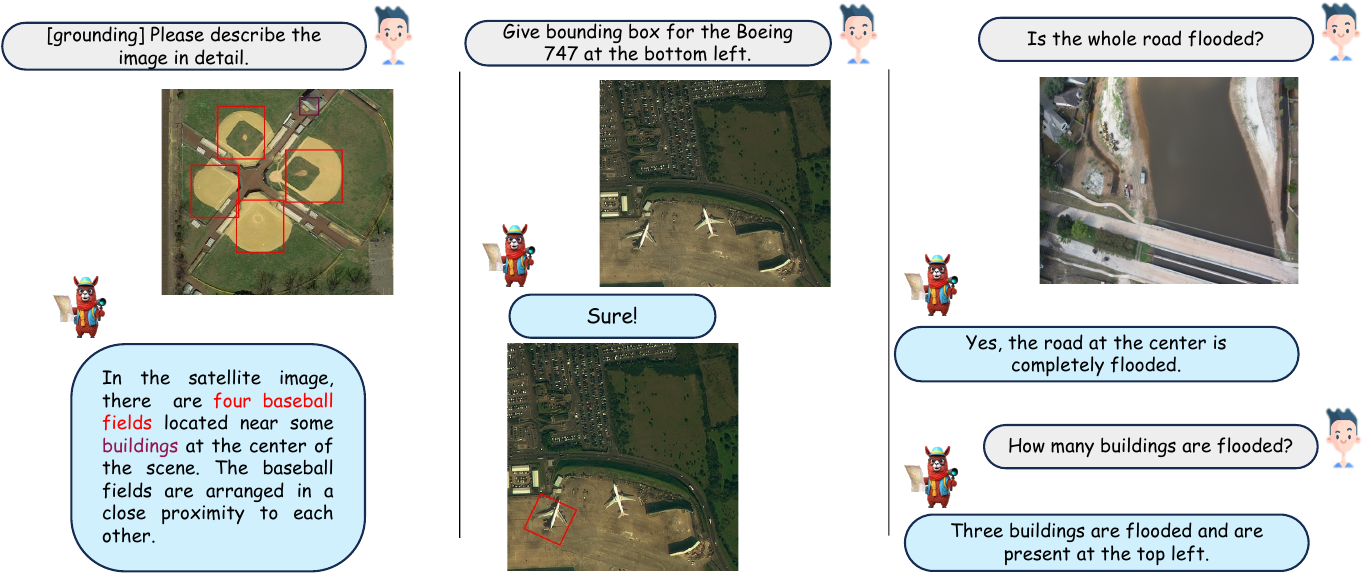}
        
    \end{center}
\vspace{-0.5cm}
    \caption{Qualitative results of GeoChat. (\emph{left-right}) Results are shown on grounding, referring object detection, and disaster/damage detection. The user can provide task-specific tokens (e.g., \texttt{[grounding]}) to shape model responses according to the desired behavior. The model can generate textual responses (\emph{right}), only visual grounding (\emph{center}) and both text and object groundings interleaved together (\emph{left}). The model can also specify object types, object counts, object attributes and object relationships. }
    \label{fig:graph}          
\end{figure*}
\section{RS Multimodal Instruction Dataset}
\label{sec:dataset_creation}

By using LLM Vicuna \cite{vicuna2023}, we align the model to follow a range of instructions by presenting and curating varied instruction-following data with multi-round conversations regarding remote sensing imagery (Table \ref{tab:instruction_following_data_mixture}). We specifically provide system instructions  as prompts that ask Vicuna \cite{vicuna2023} to generate multi-round question and answer pairs in a manner as if it could visualize the image (although it only has access to the text). This is achieved by providing few-shot in-context examples  manually composed within the prompt to show Vicuna \cite{vicuna2023} how to build high-quality instruction-response pairs based on the caption and information supplied. Specifically, from our short descriptions created using the below pipeline, we randomly sample 65k images to create multi-round conversations, 10k images to generate complex question answers and 30k images to generate detailed descriptions for the given short descriptions. 

In combination, after conversion to instruction format, we obtain a total of nearly 306k image-instruction pairs for training and 12k for testing. Next, we outline the instruction-set creation process.

\begin{table}[h]
\vspace{-0.3cm}
\resizebox{\columnwidth}{!}{
  \centering
  \begin{tabular}{llccr}
    \toprule
  \rowcolor{mygray}  Dataset & Category   & \# Classes    & \# Images & Image Size \\
    \midrule
    DOTA & Object Detection & 18  & 17,480 & 1024 $\times$ 1024    \\
    DIOR & Object Detection & 20  & 23,463 & 800 $\times$ 800  \\
    FAIR1M & Object Detection & 37  & 64,147 & 600 $\times$ 600 \\
    LRBEN(rsvqa) & Visual Question Answering & - & 600 & 256 $\times$ 256 \\
    Floodnet & Visual Question Answering & - & 4056 & 3000 $\times$ 4000 \\
    NWPU-RESISC-45 & Scene Classification & 45  & 31,500 & 256 $\times$ 256 \\
    \bottomrule
    
  \end{tabular}
  }
    \caption{List of datasets used to creat our remote-sensing instruction set for GeoChat VLM training. We include object detection, visual question answering and scene classification datasets with varying image sizes and types of classes to ensure diversity.}
    \label{tab:total_dataset}
\end{table}

\begin{table*}[t]
\begin{minipage}{1\columnwidth}
  \centering
  \scalebox{0.9}{
  \begin{tabular}{lll}
    \toprule
   \rowcolor{mygray} &Attribute     & Example \\
    \midrule
   a1 & category & (e.g. “plane, ship”)    \\
    a2 & color & (e.g. “gray, white”)    \\
    a3 & relative size & (e.g. “small, large”)    \\
    a4 & relative location & (e.g. “top right, bottom”)    \\
    a5 & relation & (e.g. “parked at, driving through”)    \\
    \bottomrule
  \end{tabular}}
  \caption{List of attributes collected for objects. Attributes are used to obtain referring expressions e.g., small-sized plane to the left.}
      \label{tab:attributes}
\end{minipage}
\hfill
\begin{minipage}{.49\textwidth}
\centering
    \setlength{\tabcolsep}{3pt}
\scalebox{0.78}[0.78]{
  \begin{tabular}{ll}
    \toprule
    \rowcolor{mygray} Categories     & Example \\
    \midrule
    Ships and Harbors & (e.g. “anchored at, parked at”)    \\
    Track Field and Soccer Field & (e.g. “Surrounded by, Inside”)    \\
    Vehicles, Bridge, Road, Roundabout &  (e.g. “passing through, passing through”)    \\
    Vehicles and Building &  (e.g. “parked”)    \\
    Airport and Plane &  (e.g. “parked”)    \\
    Ship and Helipad & (e.g. “on, contains”)    \\
    \bottomrule
  \end{tabular}}
    \caption{Example of relationships between different objects used in the proposed instruction dataset.}
        \label{tab:relations}
    \end{minipage}
    \end{table*}


\textbf{Constituent Datasets:} 
In the compilation of our instruction set, we incorporate three distinct types of datasets, encompassing the ones designed for object detection, scene classification, and visual question answering (VQA). Specifically, we integrate three object detection (DOTA \cite{Xia_2018_CVPR}, DIOR \cite{dior}, and FAIR1M \cite{FAIR1M} which together form the SAMRS \cite{SAMRS} dataset), one scene classification (NWPU-RESISC-45 \cite{cheng2017remote}), one VQA (LRBEN\cite{lobry2020rsvqa}), and one flood detection \cite{rahnemoonfar2020floodnet} VQA dataset (see Table \ref{tab:total_dataset}). The object detection datasets allow region-level reasoning capability as they offer segmentation masks along with bounding boxes.

    
    

\textbf{Addition of Missing Classes:} Although a wide variety of object classes are included in the object detection databases, several essential categories like buildings, roads, and trees are missing. To address this, we propose to utilize ViTAE-RVSA \cite{rvsa} model, pre-trained on the LoveDA dataset \cite{junjue_wang_2021_5706578}, which encompasses the required important classes. The model \cite{rvsa} is used to infer these classes on the SAMRS \cite{SAMRS} dataset, yielding pseudo labels. To mitigate potential noise in these predictions, we remove the predictions of ViTAE-RVSA \cite{rvsa} for which we already have ground truth from the SAMRS \cite{SAMRS} dataset to refine the results. 


\textbf{Attribute extraction:}
For referring expression annotations, it is important to derive a variety of attributes in RS images. 
To this end, we have selected five distinct types of attributes, as outlined in Table \ref{tab:attributes}. 
Object category information can be directly obtained from the SAMRS dataset. 
For color extraction, we use the K-Means clustering algorithm. Specifically, we extract the object's pixels from the image using ground-truth box and cluster them into $K$ groups. The center of the largest cluster is then selected as the object's color. 
To specify the relative size of the object, we categorize objects into three sizes: small, normal, and large. 
This categorization is determined by measuring the area of all instances of a class in the entire dataset and assigning the 80$^{th}$ percentile as the large label. 
Similarly, the 20$^{th}$ percentile is designated as small size, with the remaining falling into the normal category. 
To determine the object's relative position within the images, we partition the entire image into a 3$\times$3 grid, defining regions such as Top Right, Top, Top Left, Left, Center, Right, Bottom Right, Bottom Left, and Bottom. Based on the object's center pixel coordinates, we assign its relative position accordingly.

To define the relation between objects in a given image, we group different objects based on their distance between the bounding boxes, and for each sub-graph, we assign different relationships between objects based on their class labels. Table \ref{tab:relations} presents various examples of object relationships. To establish relationships like ``surrounded by," we cross-reference pixel-level coordinates to verify if one object is entirely contained within another object. 

\textbf{Expression Generation:}
To emulate natural language expressions, we employ predefined textual templates based on \cite{RSVG}. 
The phrase template encompasses the attributes \{a1, \dots, a5\} from Table \ref{tab:attributes}. The expression for a group of objects of the same class is formulated as:
\begin{align}\notag
\text{"The/A $\langle a3 \rangle$ $\langle a2 \rangle \,a1 \langle$ in/on  the  $a4 \rangle$ ." }
\end{align}
Attributes that may be absent are enclosed in ⟨⟩, and attributes \{a2, a3\} can be arranged in any sequence.

 \begin{table}[!t]
  \centering
      \setlength{\tabcolsep}{12pt}
    \scalebox{0.9}[0.9]{
  \begin{tabular}{lcccc}
    \toprule
    \rowcolor{mygray} Model & UCMerced & AID \\
    \midrule
    Qwen-VL \cite{bai2023qwen} &62.90 & 52.60 \\
    MiniGPTv2 \cite{chen2023minigpt} & 4.76 & 12.90 \\
    LLaVA-1.5 \cite{liu2023improved}  & 68.00 & 51.00\\
    \rowcolor{myblue} GeoChat & \textbf{84.43} & \textbf{72.03 }\\
    \bottomrule
  \end{tabular}}
    \caption{Zero-shot scene classification accuracy comparison on {AID} \cite{xia2017aid} and {UCMerced} \cite{yang2010bag} datasets. In comparison to other generic VLMs, GeoChat performs favorably well.}
    \label{tab:scene_classify}
\end{table}

Similarly, the sentence template incorporates the relational attributes a5 to establish connections between two objects through this structure:
\begin{align}\notag
\text{"The/A $\langle a_i3 \rangle$ $\langle a_i2 \rangle \,a_i1  \,a_i5 \,a_j1 \langle$ in/on  the  $a_j4 \rangle$."}
\end{align}
Here, the indicies \(i\) and \(j\) represent the \(i^{th}\) and \(j^{th}\) object. 

\textbf{Visual Grounding:}
Although referring expression datasets are available in the natural image domain \cite{young2014image, yu2016modeling}, they lack for the remote sensing domain. To this end, we use our short descriptions as referring expressions to create three different kinds of question answering pairs, i.e. grounding image description, referring expression, and region level captioning, as described in Table \ref{tab:instruction_following_data_mixture}.

\section{Experiments}

\subsection{Implementation Details}
We initialize the weights of our model with the pretrained CLIP-ViT \cite{radford2021learning}, and LLM (Vicuna-v1.5 \cite{vicuna2023} and 
 apply LoRA \cite{lora} finetuning. Utilizing LoRA, we refine the parameters \(W_q\) and \(W_v\) through low-rank adaptation, with a designated rank \(r\) set to 64 in our implementation. 
 The model undergoes training consistently at an image resolution of 504 $\times$ 504 throughout the whole process. 
 Each training step incorporates specifically crafted multi-modal instructional templates designed for a variety of vision-language tasks during the training process. 
 We use AdamW \cite{loshchilov2017decoupled} optimizer with a cosine learning rate scheduler to train our model. 
 We keep the global batch size as 144. 
 We train our model in two stages, first, we train using all of our datasets for 1 epoch, correspondingly 2400 steps, followed by stage 2, where we only train on the grounding dataset for 1600 more steps.

\subsection{Scene Classification}  
\textbf{Datasets for evaluation.} For scene classification, we evaluate our model using AID \cite{xia2017aid} and UCMerced 
 \cite{yang2010bag}. AID \cite{xia2017aid} is a large-scale aerial image collection compiled from Google Earth imagery, with 30 classes, such as a river, dense residential area, etc. The images are labeled by specialists in the field of remote sensing image interpretation. In total, the AID \cite{xia2017aid} dataset has 10,000 images within 30 classes. The images have been taken from different countries as well as different weather conditions. For evaluation, we use a 20\% split of the AID \cite{xia2017aid} dataset. UCMerced \cite{yang2010bag} is a Land Use scene classification dataset, with 2,100 images and 21 classes. Each image is of size 256$\times$256. We use the whole  UCMerced \cite{yang2010bag} dataset as a zero-shot test set.

\textbf{Results.} We prompt the models with all of the classes and prompt to classify the image using just one word/phrase. For example, we input a prompt like \texttt{"Classify the image within one of the given classes: dense residential area, \ldots, school. Answer with one word or short phrase."}. We calculate zero-shot accuracy on both AID and UCMerced. GeoChat significantly outperforms other VLM's with an accuracy of 84.43\% on UCMerced \cite{yang2010bag} and 72.03\% on AID \cite{xia2017aid}, as presented in Table \ref{tab:scene_classify}. Notably, the recent MiniGPT-4-v2\cite{chen2023minigpt} fails to follow the instructions provided for this specific task and returns unrelated classes that are not a part of the dataset. It's accuracy is close to 5\% if we pass the answers from Vicuna-v1.5~\cite{vicuna2023} and ask it to check if the output sentence refers to the ground truth class or not. 
In comparison, Qwen-VL and LLaVa-1.5 perform well in instruction following, but fall short to GeoChat, due to lack of domain knowledge.

\begin{table}[!t]
    \centering
    \setlength{\tabcolsep}{2pt}
    \resizebox{\columnwidth}{!}{
  \begin{tabular}{lcccc}
    \toprule
    \rowcolor{mygray} Method & Presence   & Comparison    & Rural/Urban & Avg. Accuracy \\
    \midrule
        LLaVA-1.5\cite{liu2023improved} & 55.46 & 68.20 & 59.00 & 62.77 \\
    Qwen-vl-Chat \cite{bai2023qwen} & 38.57  & 67.59 & 61.00 & 55.35 \\
    MiniGPTv2 \cite{chen2023minigpt} & 55.16 &55.22 &39.00 & 54.96\\
    \midrule
    RSVQA\cite{lobry2020rsvqa} & 87.47 & 81.50  & 90.00 & 86.32    \\
    EasyToHard\cite{yuan2022easy} & 90.66 & 87.49  & 91.67 & 89.94   \\
    Bi-Modal\cite{bazi2022bi}& 91.06 & 91.16 & 92.66 & 91.63   \\
    SHRNet \cite{zhang2023spatial} & 91.03 & 90.48 & 94.00 & 91.84  \\
    RSGPT\cite{hu2023rsgpt} & 91.17 & 91.70 & 94.00 & 92.29   \\
    \midrule
    \rowcolor{myblue} \textbf{GeoChat} & \textbf{91.09} & \textbf{90.33} & \textbf{94.00} & \textbf{90.70} \\
    \bottomrule
    \end{tabular}}
    \caption{Comparisons with general zero-shot (top) and RS-VQA specialized (middle) models on RSVQA-LRBEN \cite{lobry2020rsvqa} dataset for VQA task. \cite{bai2023qwen,chen2023minigpt,liu2023improved} are evaluated in zero-shot setting. GeoChat outperforms other zero-shot models and performs competitively to SoTA-supervised models like RSGPT which are specifically fine-tuned on target dataset (while ours is a generic model not specifically finetuned on target dataset). } 
        \label{tab:LRBEN}

\end{table}

\begin{table*}[!t]
\setlength{\tabcolsep}{4pt}
\resizebox{\textwidth}{!}{
\begin{tabular}{lcccccccc}
\toprule
\rowcolor{mygray} Model    & Small & Medium & Large & Single-object grounding & Multi-object grounding & \texttt{[refer]}  & \texttt{[grounding]}                & Overall \\ \midrule
MiniGPTv2 \cite{chen2023minigpt} & 1.7   & 9.9    & \textbf{21.9}  & 9.1    & 3.6   & 8.2  & \multicolumn{1}{c}{2.6}  & 7.6     \\
\rowcolor{myblue} GeoChat      & \textbf{2.9}   & \textbf{13.6}   & {21.7}  & \textbf{16.0}   & \textbf{4.3}   & \textbf{10.5} & \multicolumn{1}{c}{\textbf{11.8}} & \textbf{10.6}    \\ 
\bottomrule
\end{tabular}}
\caption{Performance (acc@0.5\%) comparison of GeoChat on our benchmark. Small, medium and large refer to the size of the objects based on the bounding box area. Single/multi-object refer to how many objects the question asks the model to predict. \texttt{[refer]}: object referenced using one attribute from a2, a3 or a4 in Table \ref{tab:attributes}. \texttt{[grounding]}: objects referenced using a combination of attributes from a1-a5 in Table \ref{tab:attributes}. Overall, GeoChat outperforms the baseline, but there is still significant room for further improvement on this complex task.} \label{tab:grounding_acc}
\vspace{-12pt}
\end{table*}

\begin{table}[!htp]
  \centering
      \setlength{\tabcolsep}{4pt}
    \scalebox{0.95}[0.95]{
  \begin{tabular}{lcccc}
    \toprule
    \rowcolor{mygray} Model & Presence   & Comparison  & Average Accuracy \\
    \midrule
    Qwen-VL\cite{bai2023qwen} & 66.44 & 60.41 & 63.06 \\
    LLaVA-1.5\cite{liu2023improved} & 69.83 & 67.29 & 68.40\\
    MiniGPTv2\cite{chen2023minigpt} &40.79 & 50.91 & 46.46 \\
    \rowcolor{myblue} GeoChat  & \textbf{58.45} & \textbf{83.19}  & \textbf{72.30}\\
    \bottomrule
  \end{tabular}}
    \caption{Comparison with other general ZS model's on {RSVQA-HRBEN} \cite{lobry2020rsvqa} dataset for visual qa. All models here have not been trained on the target dataset. GeoChat performs favorably well compared to generic VLMs. }
    \label{tab:HRBEN}
    \vspace{-10pt}
\end{table}


\subsection{Visual Question Answering}

\textbf{Datasets for evaluation.} RSVQA-HRBEN \cite{lobry2020rsvqa} comprises 10,569 high-resolution photos and 1,066,316 question-answer pairs, with 61.5\%, 11.2\%, 20.5\%, and 6.8\% divided into training, validation, test 1, and test 2 sets, respectively. This dataset has three question types: presence, comparison, and count. For evaluation, we use the test set-2 for RSVQA-HRBEN \cite{lobry2020rsvqa} with 47k question answer pairs. RSVQA-LR \cite{lobry2020rsvqa} is made up of 772 low-resolution images and 77,232 question-answer pairs, with 77.8\%, 11.1\%, and 11.1\% used for training, validation, and testing, respectively. There are four different categories of questions: presence, comparison, rural/urban, and count. We omitted area and count questions during evaluation because the responses are numerical and quantifiable into numerous categories. In the RSVQA-LRBEN \cite{lobry2020rsvqa} dataset, for example, counting questions are quantified into five categories: 0, between 1 and 10, between 11 and 100, between 101 and 1000, and greater than 1000. For evaluation, we use the test set of RSVQA-LRBEN \cite{lobry2020rsvqa} with 7k question-answer pairs.

\textbf{Results.} To constrain the answers to a simple yes/no and for rural/urban question types, we add a suitable prompt
at the end of each question. GeoChat performs close to the SOTA specialist models on RSVQA-LRBEN test set, which is RSGPT \cite{hu2023rsgpt}, finetuned on the target dataset for 5 iterations in comparison. We also match the SOTA on urban-rural classification subset, as presented in Table \ref{tab:LRBEN}. For RSVQA-HRBEN, GeoChat outperforms other VLM's in zero-shot setting on average accuracy by 3.9\%, while beating the Comparison subset by 15.9\% on LLaVA-v1.5 \cite{liu2023improved}, as shown in Table \ref{tab:HRBEN}.

\subsection{Visual Grounding}
\textbf{Datasets for evaluation.} For the evaluation of grounding tasks, we propose a new benchmark that contains different referring and grounding tasks. We use the validation set from \cite{SAMRS} and used the same dataset creation pipeline as in Sec.~\ref{sec:dataset_creation} to construct the test benchmark. There are a total of 7653 [refer], 758 [grounding], and 555 grounding description questions. We use accuracy@0.5 as the evaluation metric. Accuracy is calculated if the predicted box has an overlap of more than 0.5 IoU with the ground-truth box.

\begin{table}[!t]
\centering
\setlength{\tabcolsep}{8pt}
\scalebox{0.9}{
\begin{tabular}{lccc}
\toprule
\rowcolor{mygray} Model    & acc@0.5 & acc@.25 & METEOR \\ 
\midrule
MiniGPTv2\cite{chen2023minigpt}     & 10.8    & 30.9    & 16.4                        \\ 
\rowcolor{myblue} GeoChat &  \textbf{11.7}    & \textbf{33.9}    & \textbf{48.9}                        \\ 
\bottomrule
\end{tabular}}
\vspace{-5pt}

\caption{Results on grounding description task.}\label{tab:grounding_desc_acc}
\vspace{-3pt}
\end{table}

\begin{table}[!t]
\centering
\scalebox{0.9}{
\vspace{-12pt}

\begin{tabular}{lccc}
\toprule
\rowcolor{mygray} Model    & ROUGE-1 & ROUGE-L & METEOR \\ 
\midrule
MiniGPTv2\cite{chen2023minigpt} & 32.1    & 31.2    & 10.0   \\
\rowcolor{myblue} GeoChat      & \textbf{87.3}    & \textbf{87.2}    & \textbf{83.9}   \\ 
\bottomrule
\end{tabular}}
\caption{Region level captioning performance.}\label{tab:region_caption_acc}
\vspace{-10pt}

\end{table}

\textbf{Results.} Table~\ref{tab:grounding_acc} shows the performance of our method and MiniGPT-4-v2 \cite{chen2023minigpt} on the proposed benchmark. Overall, the model performance is low on small objects or when it has to predict multiple boxes. Compared to MiniGPT-4-v2\cite{chen2023minigpt}, our model works better on medium size images. On the grounding description task,
we calculate both, the IoU for the multiple bounding boxes generated 
 as well as the text answer generated. Our model provides a better description with slightly better box accuracy than MiniGPT-4-v2 \cite{chen2023minigpt} (Table \ref{tab:grounding_desc_acc}). As for region-level captioning, we evaluate both models based on the text accuracy with ground truth region-level captions (Table \ref{tab:region_caption_acc}). Our model significantly outperforms MiniGPT-4-v2 in terms of ROUGE and METEOR score.

\section{Conclusion}
Although recent advancements in large Vision-Language Models (VLMs) have shown promise in nature image domains, their performance in Remote Sensing (RS) scenarios is still limited due to the unique domain-specific challenges. Addressing this gap, we present GeoChat, the first unified remote sensing VLM that excels in multitask conversational capabilities with high-resolution RS images. GeoChat not only answers image-level queries but also engages in region-specific dialogue, grounding responses with precise spatial coordinates. We create a novel RS multimodal instruction-following dataset comprising of $318k$ image-instruction pairs with a diverse multitask format. GeoChat achieves robust zero-shot performance across various RS tasks including scene classification, VQA, multi-turn dialogue, visual grounding and referring object detection, thus establishing a comprehensive benchmark.

    

{
    \small
    \bibliographystyle{ieeenat_fullname}
    \bibliography{main}
}


\end{document}